# Graph Neural Network based Agent in Google Research Football

Jinglong Liu[1, †]
[1]College of Computer and Information Engineering
Central South University of Forestry and Technology
Changsha, China
ljl894327911@outlook.com

Yizhan Niu[2, *, †]
[2]Department of Computer and Information Science
Fordham University
New York, NY 10023, USA
*yniu9@fordham.edu

Yuhao Shi[3, *, †]
[3]College of Computer Science and Technologies
Zhejiang University, Hangzhou, China
*3180105072@zju.edu.cn

Jiren Zhu[4, †]
[4]School of International Studies
Zhejiang University
Hangzhou, China
3170106340@zju.edu.cn

[†]These authors contributed equally.

*Abstract*--**Deep neural networks (DNN) can approximate value functions or policies for reinforcement learning, which makes the reinforcement learning algorithms more powerful. However, some DNNs, such as convolutional neural networks (CNN), cannot extract enough information or take too long to obtain enough features from the inputs under specific circumstances of reinforcement learning. For example, the input data of Google Research Football, a reinforcement learning environment which trains agents to play football, is the small map of players' locations. The information is contained not only in the coordinates of players, but also in the relationships between different players. CNNs can neither extract enough information nor take too long to train. To address this issue, this paper proposes a deep q-learning network (DQN) with a graph neural network (GNN) as its model. The GNN transforms the input data into a graph which better represents the football players' locations so that it extracts more information of the interactions between different players. With two GNNs to approximate its local and target value functions, this DQN allows players to learn from their experience by using value functions to see the prospective value of each intended action. The proposed model demostrated the power of GNN in the football game by outperforming other DRL models with significantly fewer steps.**

## I. Introduction

Games, as abstractions of real-world problems, are often seen as benchmarks for reinforcement learning algorithms. As a branch of Artificial Intelligence, reinforcement learning improves an agent's performance through trial and error for specific tasks that require human intelligence. With the increasing success cases of reinforcement learning in playing games, researchers now have more confidence in AI's capacity to solve real-world decision-making problems. Current studies that attempt to improve reinforcement learning results (e.g., faster convergence and higher scores) focus on different aspects, such as a more efficient deep learning model to approximate functions, a more complex reinforcement learning algorithm, and a simulated environment that is closer to the real-world scenario [1] [2] [3] [4] [5] [6]. This paper aims to develop the algorithm that uses a GNN to train agents to play football against built-in bots in Google Research Football Environment.

To address the problem that it is difficult to find an exact function that represents either a value function or a policy for a reinforcement learning problem, researchers introduce the methods of deep learning as a way to approximate different functions. In 2015, researchers at DeepMind Technologies introduced the pioneering algorithm DQN, which integrates deep learning methods with reinforcement learning algorithms [1]. They used a CNN to approximate the value function of each observation to inform the agents which action is more valuable to do in each situation. The first combination did exceed previous AI methods in playing pixel video games, but this model could hardly complete more complex tasks. Two years later, AlphaGo becomes another focus of the reinforcement learning world [3]. AlphaGo uses a CNN and fully connected layers to convert a game board input into a probability distribution indicating how a player should act. This AI agent defeated Jie Ke, who was a top GO player. Although AlphaGo achieved superhuman performance in the GO game, the training complexity of this considerable model went far beyond average researchers' capacity: it was trained with 64 GPU workers and 19 CPU parameter servers. Later on, researchers in Google tried combining sequential models, including RNN and LSTM with RL methods to extend the length of memories and update the gradients with hundreds of steps backward [7] [8]. At the same time, Kulkarni, Tejas D., et al. proposed the integration of DQN and Hierarchical Reinforcement Learning (HRL), which is an improvement of DRL on the RL algorithm part [9]. It is worth mentioning that the football environment developers, on which environment this proposed paper is based, also utilized the method of DRL [10]. They tried three different algorithms including IMPALA, PPO, and DQN [5] [6] [11]. They all did a relatively good job after 500 million training steps, while such a big number of training steps take a massive amount of computational time to obtain some proportion of the information from the inputs.

These DRL algorithms work relatively well after training for long time to extract the information from inputs. However, these models either can hardly obtain much information or take too long to train in this specific problem setting of training football agents. Addressing this problem, the proposed algorithm uses a GNN as the model in a DQN algorithm. Choosing a GNN, especially a GCN, allows the proposed algorithm to have a strong ability of information transferability between the graph filters [12]. Finally, using the GNN-based DQN model, under certain conditions, the trained football team can achieve satisfactory results in different modes of football games within fewer training steps.

The proposed method uses a DQN algorithm with a GNN as the model. The raw information of the football game is converted into a graph with players' states as its node features. By doing so, this model can reduce two problems within the previous methods. First, under certain circumstances, a GCN can better transmit information than a CNN when representing relationships between data [13]. Second, the pixel input the previous algorithms used in the football environment was a matrix with many pixel values which contained no information. This model can extract the features from observations to the greatest extent by transforming the raw information into a graph. Since the location of each football player forms like a graph, the graph representation in this model is also assumed to be able to simulate real world information transmission.

The rest of this paper is organized as follows: Section 2 introduces the fundamental knowledge of reinforcement learning environments, graph convolutional networks (GCN), and GCN-based reinforcement learning, as well as related works in these fields. Section 3 discusses the methods such as the structure of the proposed model. Section 4 displays the results of experiments in different modes of football games. Finally, Section 5 will conclude this paper and discuss the future works.

## II. RELATED WORKS

There have already been practices focusing on reinforcement learning environment engine and models which take advantages of Graph Neural Network (GNN). Our work is derived from works about Graph Attention Networks (GAT) which is one kind of GNN and works about game environment engine.

**Reinforcement Learning Environment** Earlier works have been looking into traditional convolution network in reinforcement learning. The gridworld [11] took the world as a matrix-like grid. It also had multi-agent version [14]. Games were also aspects that RLEs had interests in. Agent training in Atari games [15] reached super-human level. Later works focused on more complicated games such as the StarCraft [16], Minecraft [17], and Doom [18]. However, most of the environment above did not provide multi-agent training. The Arcade, Minecraft, and Doom environments only discussed single-agent games, and the StarCraft only supported two agents. The real-world was not deterministic which motivated the need to develop algorithms that can cope with and learn from stochastic environments. The environment provided us with stochastic conditions, and the Google Research Football [7] had a random space based on 11 vs. 11 players. The Google Research Football also offered open-source licenses and required no closed-source binary such as StarCraft. Open-source licenses enabled researchers to inspect the underlying game code and modify environments to test new research ideas.

**Graph Neural Networks** Nowadays, the GNN used in most experiment was derived from those described in the following papers: the review paper of GCN including the summary of early GCN methods in [19]; the summary of how to use GNNs and GCNs for relational reasoning using a unified framework called graph networks in [20]; reviews of different kinds of deep learning methods applied to graphs in [21]. GCNs were firstly invented to deal with semi-supervised classification tasks [22]. Lots of learning tasks required dealing with graph data which contains rich relation information among elements. In other domains such as learning from non-structural data like texts and images, reasoning on extracted structures, like the dependency tree of sentences and the scene graph of images, was an important research topic which also needed graph reasoning models. In recent years, systems based on graph convolutional network (GCN) have demonstrated ground-breaking performance on many aforementioned tasks [5].

**Applying GCN in Reinforcement Learning** GCN have already been applied in reinforcement learning, but the scope of its utility was relatively narrowed. The DeepPocket [8] exploited the time-varying interrelations between financial instruments with GCN, which focused on the relationships. Other usages focused on recommendation system. Works in [4] explored self-supervised learning on user-item graph, to improve the accuracy and robustness of GCNs for recommendation. There were also works done to explore STGCN in search system. In [23] they discussed about making use of existing models to automatically explore high-performance STGCN model for specific scenarios. Graph attention was also included in [3]. However, applying GCN in agent training in games such as football was seldom explored before.

Our work talks about the usage of GCN on reinforcement learning in games. We combine the flexibility of GCN with well-developed reinforcement learning environments, introducing a new network architecture. Our methods provide a method to integrate the GCN into reinforcement learning environments. Also, our method refers to Q-learning Function which is introduced in Deep Q-Network [24].

## III. METHODOLOGY

This paper presents a combination of GNNs and deep q-learning algorithm to train the agent to play the football game presented by google. This section illustrates details about using the provided *gfootball* environment, the structure of the network to represent the observations of the game, and the algorithm used to train the network. The whole process can be represented by the Fig.1 below:

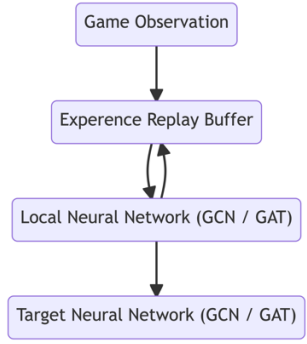

Figure 1. The Project Flowchart

### A. Gaming Environment

- **Goal**: As mentioned above, the proposed method aims to train agents to play football on a specific environment Google Research Football, also known as gfootball, an open-source AI research platform on which researchers can train their football RL agents.
- **Observation**: Each game under a specific environment returns the observation of every game situation, which is a dictionary of all the information of the ball and the players on both sides.
- **Scenario**: The environment provides various scenarios that can be used in training the agent under different circumstances such as counterattack and 11-vs-11 competitions.
- **Reward**: The environment provides two rewards, scoring and checkpoints. Scoring is the default reward mode, which gives the reward to agents when they score in the competition. Checkpoints is another type of rewards, which checks the possession of the ball. If the agents can dribble the ball to a reward region, they will be rewarded. The proposed method chooses scoring and checkpoint rewards to better simulate a real-world scenario.
- **Evaluation**: Results of the agent are reflected by the matching results of the agent's game against the built-in rule-based game AI. The score difference between both teams will be the critical standard to evaluate the effectiveness of the training algorithm.

### B. Algorithm Details

- **Environment**: a built-in 11-vs-11 environment scenario with the scoring, checkpoints as its rewards
- **Action_size**: dimension of the actions in the game, obtained from the environment.
- **State_size**: dimension of the observations in the game, obtained from the environment.
- **Q_network_local**: approximated value function with local GCN or GAT networks, trained directly in every step with the training data from the experience replay buffer.
- **Q_network_target**: approximated value function with target GCN or GAT networks, updated from the local network every several steps.
- **Q_network_optimizer**: optimizer provided by PyTorch, its optimizing parameters being the ones from the local network.
- **Memory**: the experience replay buffer, to save experiences and use them as training datasets.

### C. Model Definition

A football game contains various information of the players of both team as well as the ball location. Since the positions of the ball and players naturally form a graph, it is suitable to represent the input information as a graph structure. This paper tries to change the traditional CNN as a model of the DQN algorithm and will use two different types of GNNs to represent the game to obtain better performances.

The proposed method specifically chooses to use GCN and GAT, which are two powerful GNN variants. The structures of both GNNs are shown below:

*1) GCN Network*

A two-layered GCN network is employed to represent the observations from the game (Fig.2). There are 9 features in the first layer representing the features of each graph node, 32 features between layer 1 and layer 2, and 19 output numbers as the approximated values of actions. The proposed algorithm can approximate the value function with the GCN in the DQN algorithm.

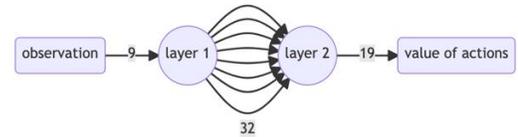

Figure 2. The GCN Network

*2) GAT Network*

As an alternative to GCN, a GAT network was also implemented to estimate the values of each action taken by the football agent. The GAT network was first published by Veličković et.al [25] in 2017. Based on the architecture described in this paper, the proposed method created a network by adding a linear fully connected layer that computes attention coefficients according to Equation (1):

$$e_{ij} = a(\mathbf{W}\vec{h_i}, \mathbf{W}\vec{h_j}) \quad (1)$$

In addition, a softmax function to normalize the coefficients across the nodes according to Equation (2):

$$\alpha_{ij} = \text{softmax}_j(e_{ij}) = \frac{\exp(e_{ij})}{\sum_{k \in N_i} \exp(e_{ik})} \quad (2)$$

Here $N_i$ is the neighborhood node of node $i$. In this paper two of the GAT layers were used as well to create the GAT network.

### D. Training Algorithm

A deep-Q learning algorithm with experience replay buffer is employed to train the network, the algorithm was first published by Mnih et.al in 2013 [1]. This method starts first by randomly playing the game and calculates the values of each action. The record of each play includes the state, the action, the reward, and the next state. The information of this play record is saved into the experience replay buffer for further training and serves as the experience of the RL agent. The agents will either take an action randomly with a probability, following the epsilon-greedy policy, or choose an action based on the value function approximated with the records of previous actions and states. Also, the local value function is updated according to the Temporal Difference (TD) loss between local and target functions. The target value function is copied from the local value function to keep the stability of the model [26].

The algorithm framework is shown in Fig.3 below:

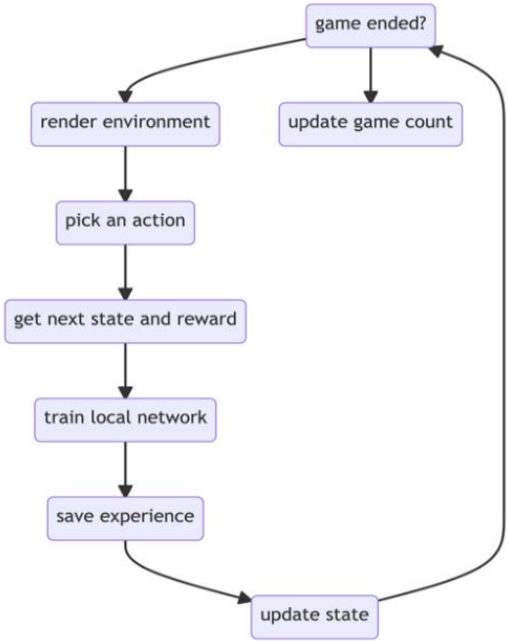

Figure 3. The DQN Algorithm

Within the DQN framework, this paper calculated the loss by taking the TD loss between two approximated value functions, the local value function and the target value function. TD local is an array of estimated action values by a local value network, which is an estimation of the current state. TD target is an array of estimated action values by a target value network, which estimates the current state based on the expectation of future states. The DQN algorithm updates the local value function with the difference between TD local and TD target as the loss and copies the parameters of the local function to the target function every few steps. Based on this updating rule, the result of the algorithm will be improved stably [1] [26].

The expressions of TD local and TD target are shown below by Equation (3) and Equation (4):

TD local :

$$\text{TD Local} = Q(s, a; \theta) \quad (3)$$

and TD target is:

$$\text{TD Target} = r + \gamma \max_{a'} Q(s', a'; \theta^-) \quad (4)$$

where $Q()$ is represents the value function, $s$ denotes the state, $a$ is the action, $r$ is the reward, $\gamma$ is the discount factor.

Also, the loss is decided according to Equation (5):

$$\text{Loss} = r + \gamma \max_{a'} Q(s', a'; \theta^-) - Q(s, a; \theta) \quad (5)$$

With all the elements needed, the proposed method can be adjusted to be applied to the football game in the *gfootball* environment. The input, as discussed before, is a dictionary of the information of players and the ball, which is transformed into a graph with 24 nodes: 22 nodes for players on both sides, 1 node for the ball, and 1 node for any other information, where each node has 9 features. Then, this graph is put into a GCN or GAT model, for which the details were discussed above, and transformed into an array of 19 elements, each of which represents the value of taking one specific action. Then, with the GNN models to approximate the value functions, the inputs can be converted into the estimated values of each action. Finally, the agent can be tried with the DQN algorithm proposed with a GNN as its model of value functions.

## IV. RESULTS AND DISCUSSION

### A. Experimental Environment

For the experimental environment, this paper will illustrate it in two parts: hardware and parameter settings.

For the first part, to have a fast speed of training, we use a graphics server with Intel E5-2678 and Nvidia Tesla V100 as the training platform, the graphic card is the best solution we have, it has a big video memory and an acceptable price for students, the detailed information is inferred by TABLE I. which you can check it below.

TABLE I. HARDWARE INFORMATION

| Hardware | Name |
| --- | --- |
| CPU | Intel E5-2678 v3 |
| GPU | NVIDIA Tesla V100 |
| RAM | 40G |
| Video Memory | 32G |

Additionally, our algorithm is based on Pytorch and Deep Graph Library two python packages, and in order to have a stabilized training environment, we select Ubuntu as operating system because it is a platform widely used in artificial intelligence research. Furthermore, CUDA and CUDNN are common combination to train and accelerate the process. the gfootball package is a football game platform to train AI to play soccer type game. The Detailed operating environment of our algorithm is shown in TABLE II.

From the TABLE II, the second part is mainly about the parameters we set in our algorithm, and the parameters of two different models are the same. They are episodes, learning rate, batch size and gradient. In this research, we set an epoch equals 3000 episodes, and the number of epochs we set is 100, and the learning rate is 0.0001, the batch size we set is 1, to prevent the unsteadiness of learning, the gradient is set as 0.7.

TABLE II. SOFTWARE VERSIONS

| System and Package | Version |
| --- | --- |
| Ubuntu | 16.04 |
| Python | 3.6 |
| Pytorch | 1.8.1 |
| CUDA | 11.2 |
| CUDNN | 8 |
| gfootball | 2.6 |

### B. Experiment Results

In this research, we mainly trained agent in *11 vs 11 easy stochastic*, *11 vs 11 hard stochastic*. *11 vs 11 competition* and *11 vs 11 Kaggle* these four schemata. The experiment results of both models are illustrated by Fig. 4-7.

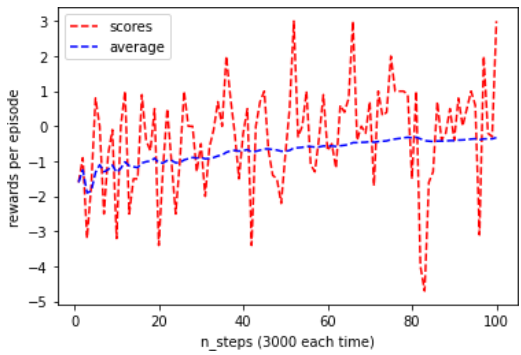

(a) GCN average score

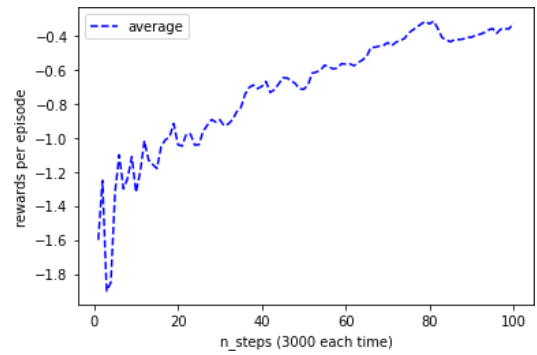

(b) GCN reward variation

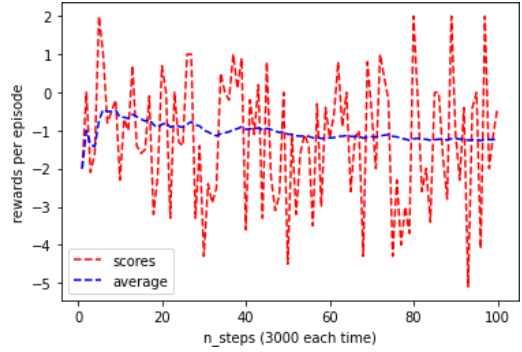

(c) GAT average score

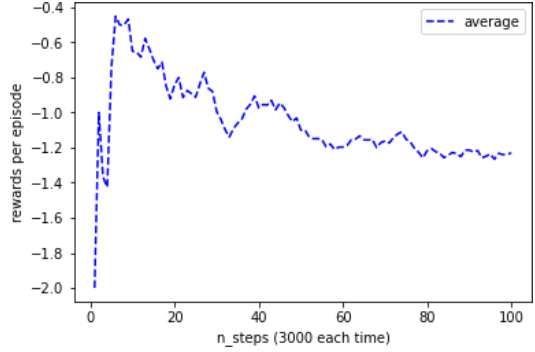

(d) GAT reward variation

Figure 4. Training Result in 11 vs 11 Easy Stochastic

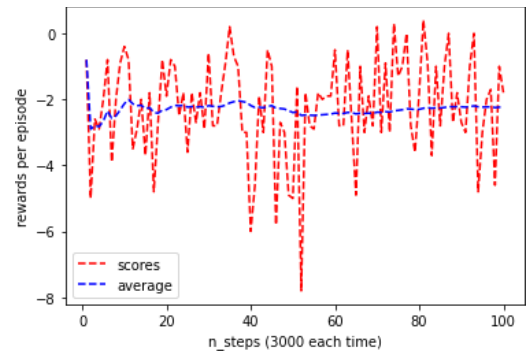

(a) GCN average score

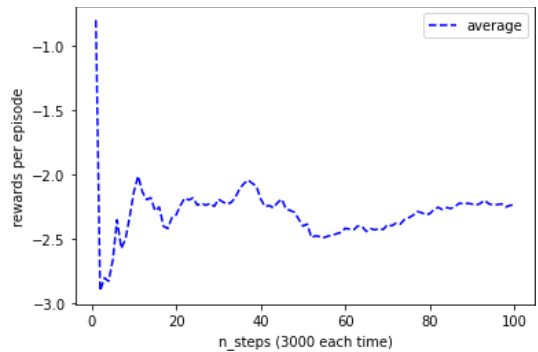

(b) GCN reward variation

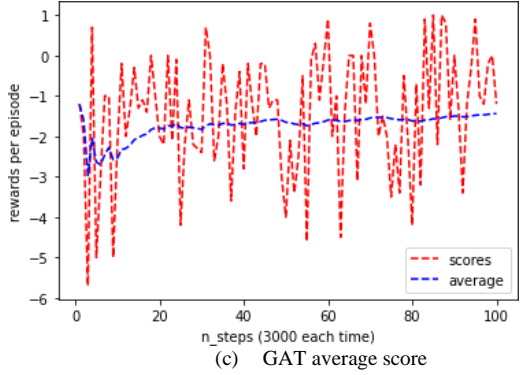
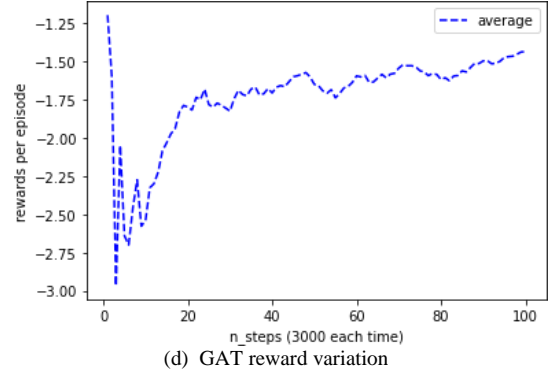

(c) GAT average score

(d) GAT reward variation

Figure 5. Training Result in 11 vs 11 Hard Stochastic

In easy schema (Fig. 4), the parameter of difficulty is 0.05, and we can find that the variation of average score of GCN is rising and finally stopped at -0.37, that is a great result in this model. Without the limitation of hardware, and increasing episodes, the variation of GCN will be better. The situation of GAT is different, the peak of variation of average score is about -0.43 at about NO.10 epoch, then continue to decline to -1.13. The reason for that maybe is the game schema is too easy to let GAT be well trained.

In hard schema, the parameter of difficulty is 0.95. The information we can be acknowledged form Fig. 5 is that the average score of GCN fluctuate in the whole training process and result is -2.23, from Fig. 5(b) we find that the variation tendency of GCN is not clear, the reason for that is our limitation of environment cannot support a longer training period, if the training period can be longer, the GCN may have a better result. Unlike the GCN, GAT have a better performance. According to Fig. 5(c) and Fig. 5(d), the variation of average score and reward are both steadily rising, and the result is -1.41. The result of GAT is excellent in just 100 epochs of training.

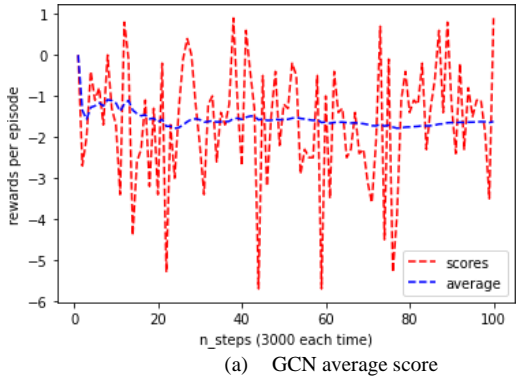
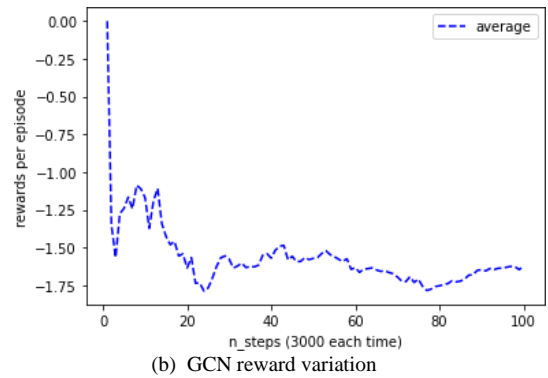

(a) GCN average score

(b) GCN reward variation

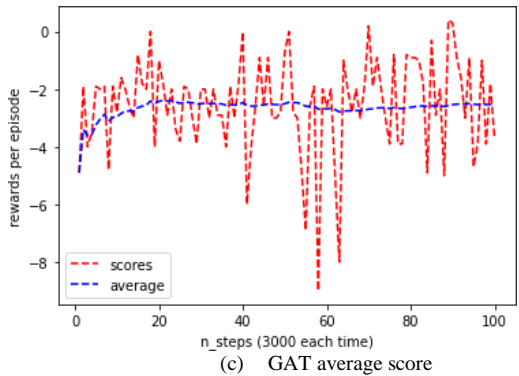
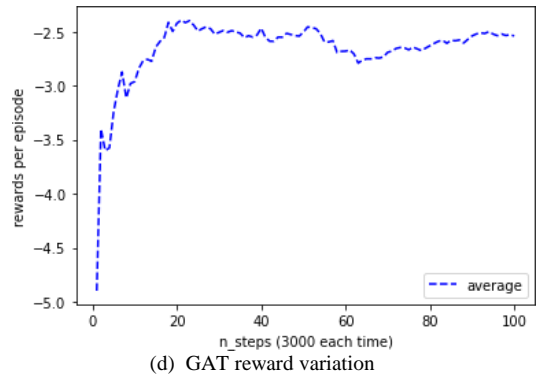

(c) GAT average score

(d) GAT reward variation

Figure 6. Training Result in Competition Schema

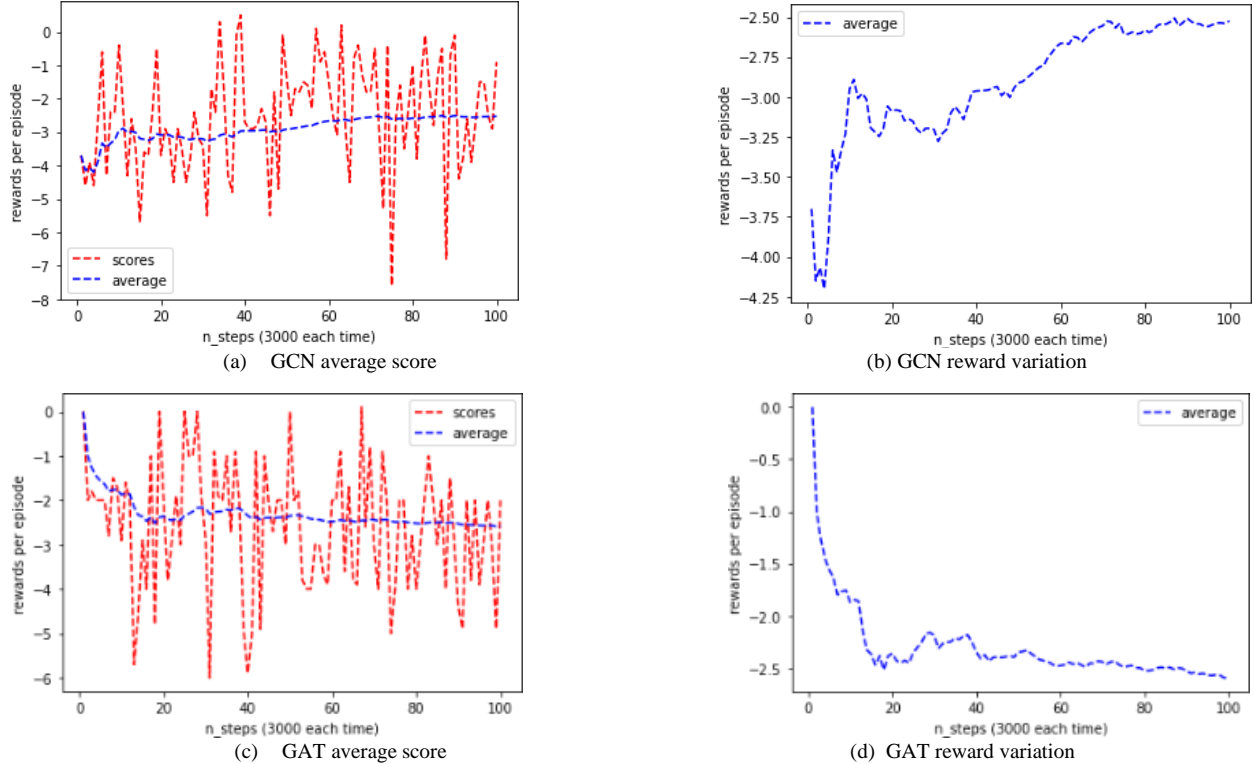

Figure 7. Training Result in Kaggle Schema

In competition schema, the computer will control two teams to have halftime games and its parameter of it is 0.6. At first, the result of GCN is -1.62, from Fig. 6(a) and Fig. 6(b), the variation of average score and reward are familiar with it in hard schema, because of the lack of training period, GCN cannot have an ideal result. Secondly, the result of GAT from Fig. 6(c) is -2.62, the result is not as good as GCN, but form Fig. 6(d), with the extension of training period, the result will be better.

The Kaggle schema is for the Kaggle Competition, the difficulty parameter is 1.0, it is the hardest schema for agent. GCN in this schema is quite well, the result is -2.52 and the variation of reward from Fig. 7(b) increased steadily, it means the result of GCN will be better and better. However, GAT in this schema is kind of disappointing, the result from Fig. 7(c) is -2.59 and the reward from Fig. 7(d) keep declining, maybe this schema is too hard for agent to have an ideal result in such short period.

From the variation of reward in different schema, the GCN have a better result in *11 vs 11 easy stochastic*,*11 vs 11 competition* and *11 vs 11 Kaggle*, but after we check on the figure of reward variation, GCN can just lead in easy schema, only in that the variation of reward and result are both increasing. In competition and Kaggle schema, the variation of GCN is not clear, so nobody can assure that with the increasing of training period, GCN could keep the lead. For GAT, the result in *11 vs 11 hard stochastic* is excellent, the training result is quite impressive, but in *11 vs 11 easy stochastic* and *11 vs 11 Kaggle*, the result is less than satisfactory, since both parameters of algorithms in the same, we consider the difficulty may cause the problem. In general, the GCN has acceptable result in both schemas, and GAT is good at hard schema. Considering of the difference of these schema, the GCN is more suitable in football games. In an experimental environment like only training in 0.3 million steps, all these results are great and can prove the success of our experiment.

C. *Comparison with Other Methods*

In former research, researchers used PPO, IMPALA and DQN to train agents in the same environment as we use, and they have done a great job in agent training [7]. Therefore, we can compare these methods in a same number of epochs, the result in hard and easy schema is shown by TABLE III. below:

TABLE III. RESULTS OF DIFFERENT MODELS IN HARD AND EASY SCHEMA

| Models | Easy | Hard | Steps |
|---|---|---|---|
| PPO | 0.05 | -1.32 | 20M |
| IMPALA | -0.01 | -1.38 | 20M |
| DQN | -1.17 | -2.12 | 20M |
| GCN | -0.37 | -2.23 | 0.3M |
| GAT | -1.21 | -1.41 | 0.3M |

From the contrast, we can find that in hard schema, the result of GAT transcends DQN and approximates IMPALA,

the result of GCN approximates DQN, the difference is quite low. In easy schema, the result of GCN transcends DQN and have gap with IMPALA, the result of GAT approximates DQN. Generally, our experiment results have a little gap with the professional one with lack of 19 million steps, as a result, GCN has a stronger learning ability in every schema, but it needs enough number of epochs to train to achieve acceptable results this way, meanwhile, GAT can attain an acceptable result in hard and competition schemas, but not good enough in other schemas. The result is quite impressive, and the experiment is successful.

## V. Conclusion

In this paper, we use GCN and GAT on Google Football platform and combine with DQN to train agent to play football games in different difficulties to proof the algorithms have a better performance in soccer video games. Therefore, based on our experimental conditions, we conduct experiments and compare results with other researchers'. The results show our approaches are efficient in low number of periods.